\documentclass{article}
\usepackage{spconf,amsmath,graphicx}
\usepackage{todonotes}
\usepackage{float}
\usepackage{enumitem}
\setlist{nosep, leftmargin=14pt}

\usepackage{mwe} 

\title{Confidence Matters: Uncertainty Quantification and Precision Assessment of Deep Learning-based CMR Biomarker Estimates Using Scan-rescan Data}
\name{\shortstack{
Dewmini Hasara Wickremasinghe$^{1}$, Michelle Gibogwe$^{1}$, Andrew Bell$^{1}$,  Esther Puyol-Ant\'on$^{1}$, \\
Muhummad Sohaib Nazir$^{1,2}$, Reza Razavi$^{1}$, Bruno Paun$^{3}$, Paul Aljabar$^{3}$, ~Andrew P. King$^{1}$ 
}}
\address{$^{1}$ School of Biomedical Engineering \& Imaging Sciences, King's College London, London, UK \\
$^{2}$ Cardio-Oncology Centre of Excellence, Royal Brompton Hospitals, London, United Kingdom \\
$^{3}$ Perspectum Ltd., Oxford, UK}
\begin{document}
%
\maketitle
\begin{abstract}
The performance of deep learning (DL) methods for the analysis of cine cardiovascular magnetic resonance (CMR) is typically assessed in terms of accuracy, overlooking precision. In this work, uncertainty estimation techniques, namely deep ensemble, test-time augmentation, and Monte Carlo dropout, are applied to a state-of-the-art DL pipeline for cardiac functional biomarker estimation, and new distribution-based metrics are proposed for the assessment of biomarker precision. The model achieved high accuracy (average Dice 87\%) and point estimate precision on two external validation scan-rescan CMR datasets. However, distribution-based metrics showed that the overlap between scan/rescan confidence intervals was ${>}50\%$ in less than 45\% of the cases. Statistical similarity tests between scan and rescan biomarkers also resulted in significant differences for over 65\% of the cases. We conclude that, while point estimate metrics might suggest good performance, distributional analyses reveal lower precision, highlighting the need to use more representative metrics to assess scan-rescan agreement.
\end{abstract}
\section{Introduction}
\label{sec:intro}

Heart function and structure can be assessed through the analysis of cardiovascular magnetic resonance (CMR) data. The interpretation of CMR data allows the estimation of functional biomarkers, which are essential to determine the health status of patients. CMR data also plays a crucial role in longitudinal analysis, as the lack of ionising radiation together with the high spatial and temporal resolution make it the imaging modality of choice to evaluate the evolution of patient biomarkers over time \cite{peterzan2016role, von2016representation}. 

The analysis of short axis (SAX) cine CMR data to estimate biomarkers consists of outlining structures of interest such as the left ventricular (LV) blood pool (LVBP), right ventricular blood pool (RVBP) and LV myocardium. This process is time consuming and prone to both intra- and inter-observer variability \cite{marwick2018ejection}. Therefore, many studies have focused on automating the process using deep learning (DL) methods \cite{bai2018automated,mariscal2023artificial,ruijsink2020fully}. These studies have shown that  DL networks are able to achieve performance comparable to humans at significantly faster rates. 

A limitation of the evaluation of DL methods for the analysis of cine CMR data is that  performance is typically assessed against manual ground truths in terms of \emph{accuracy} only, through the computation of segmentation metrics such as Dice score and/or errors in derived biomarker estimates.
However, in longitudinal analysis of CMR data the \emph{precision} of biomarker estimates is also important.
A small number of studies have evaluated the precision of DL-based biomarker estimation through the analysis of scan-rescan cine CMR datasets, 
but these have all assessed precision with respect to point estimates of biomarkers, i.e. a single biomarker value per scan \cite{bhuva2019multicenter, davies2022precision,wickremasinghe2024scan}. In reality, biomarker estimates, made by machines or humans, are subject to uncertainty. Quantifying this uncertainty offers important information for clinical purposes, particularly in longitudinal studies, but this has hitherto been neglected in research into DL-based biomarker estimation from cine CMR.

Uncertainty in DL model prediction can originate from the model (\textit{epistemic} uncertainty) or the input data (\textit{aleatoric} uncertainty). In the literature, the most common ways to estimate epistemic  uncertainty include Bayesian Neural Network methods and Monte Carlo (MC) dropout, whilst aleatoric uncertainty can be estimated using test-time augmentation (TTA). Ensemble-based methods are often used to evaluate both types of uncertainty \cite{lambert2024trustworthy}.

The work described in this paper employs both aleatoric and epistemic uncertainty estimation techniques to obtain biomarker distribution estimates, and uses these to assess biomarker precision in scan-rescan CMR datasets. The contributions of this paper are:
\begin{itemize}
\item We propose uncertainty-based DL techniques for estimating biomarker \emph{distributions} (rather than point estimates) from cine CMR data.
\item We evaluate these techniques on two scan-rescan cine CMR external validation sets.
\item We propose new metrics for quantifying precision in the context of uncertainty-based biomarker estimation.
\end{itemize}

\label{sec:intro}

\section{Methods}
\begin{figure*}[t]
    \centering
    \includegraphics[width=0.9\textwidth]{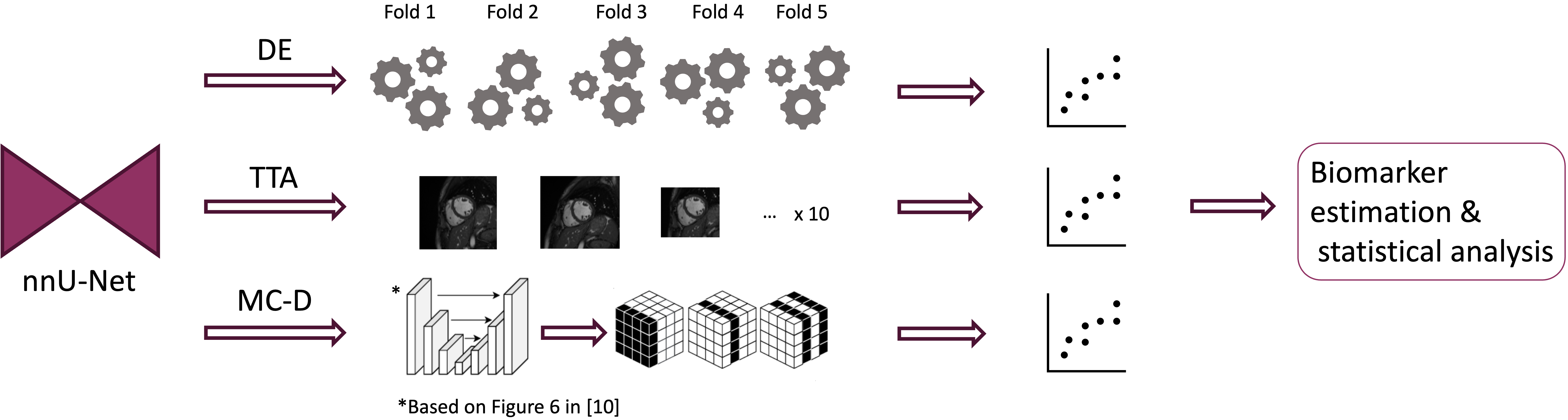}
    \caption{Overview of the proposed work}
    \label{fig:overview}
\end{figure*}

This paper evaluates the use of three different uncertainty estimation techniques, namely deep ensemble (DE), TTA, and MC dropout, in a DL-based SAX cine CMR segmentation and biomarker estimation pipeline to enable quantification of biomarker precision. Distributions of cardiac biomarkers, namely LVEF, RVEF, and LV mass (LVM), are derived from the outputs of these methods, which are then used to compute confidence limits for the biomarkers and assess the precision of the estimates on two external validation scan-rescan cine CMR datasets. An overview of the proposed pipeline can be seen in Figure \ref{fig:overview}.

\subsection{Segmentation model}
For the baseline segmentation model we employed the state-of-the-art DL model described in \cite{mariscal2023artificial}. This is based on an nnU-Net architecture \cite{isensee2021nnu} and was trained to segment the LVBP, RVBP, and LV myocardium from SAX cine CMR data. The model was trained using an adaptive loss \cite{mariscal2023artificial}, to account for labelling inconsistencies, on a database of 2793 SAX cine CMR clinical scans from the Guy’s and St Thomas’ NHS Foundation Trust (GSTT). The model was retrained following the training protocol described in the original paper, adding a dropout layer after each convolution in the encoding path, with a dropout rate of 0.2. 

\subsection{Biomarker estimation}

Each output segmentation was used to produce estimates for LVEF, RVEF, and LVM. The first step to calculate LVEF and RVEF involved the computation of the end-diastole (ED) and end-systole (ES) LV/RV volumes. The ED/ES frames had been previously manually identified in the datasets. The SAX images from these frames were input into the segmentation model and the volumes calculated by summing the foreground voxels over the SAX stack and multiplying by the voxel size. Once the volumes at ED and ES (EDV and ESV respectively) were calculated, EF values were computed as follows: $EF = \frac{(EDV-ESV)}{EDV}*100$. LVM was obtained by multiplying the volume of the LV myocardium at ED by the density of the myocardial muscle (i.e., 1.05 $g/mL$ \cite{ bai2018automated}).

\subsection{Uncertainty estimation techniques}
The first uncertainty estimation method, DE, exploits the five-fold cross validation embedded in the nnU-Net framework. Specifically, the training of an nnU-Net model involves training five separate models using cross validation, which are generally ensembled at inference time. In this work, inference was run on each separate model, to obtain five different predictions. TTA was applied on the fly at inference time. Spatial and intensity transformations were applied, including rotations (random angles within $\pm$10\textdegree), cropping (random crops up to 10\% of the total volume), Gaussian blurring (random value of $\sigma$ between 0 and 1), and intensity scaling (random $\pm$10\%). For this method ten different random augmentations were generated, resulting in ten different biomarker estimates. The MC dropout segmentation outputs were obtained by running inference while keeping the dropout layers active. This was done ten times, resulting in ten different biomarker estimates. TTA and MC dropout were applied to each cross-validation fold separately, and the final outputs of each run were ensembled as per the standard nnU-Net inference protocol.

\section{Experiments and results}

\subsection{External validation datasets}
Two external scan-rescan datasets were employed for the precision analysis. The first one, which is referred to as Roca-HC and was first described in \cite{roca2023cardiac}, was a collection of 184 scan-rescan cine CMR images acquired from 92 healthy volunteers. The second one, referred to as Strain-8 \cite{bell2025strain}, consisted of 40 scan-rescan cine CMR scans from 20 healthy volunteers acquired from the GSTT. For both datasets, the two scans belonging to the same patient were obtained on the same day, with replanning carried out before each acquisition. 


\subsection{Evaluation}
First, the mean and standard deviation of the differences between scan and rescan mean biomarker values were computed, as well as the coefficients of variation (CoV) between scans A and B corresponding to the same patient, consistently with precision analyses found in the literature. Next, we propose three new metrics for quantifying the precision of uncertainty-based biomarker estimates in scan-rescan data. In the first metric, we compute the percentage of scan B values which lie within the 95\% confidence limits of the scan A distribution, and vice versa. We refer to this metric as Confidence Precision Percentage (CPP). We also propose a metric based on the percentage overlap between the confidence limits for scan A and scan B. This is computed by dividing the intersection of the two limits by their union. We refer to this metric as Confidence Intersection over Union (CIoU).
Finally, the similarity between scan and rescan distributions for each biomarker was assessed using paired t-tests and Wilcoxon signed-rank tests, depending on the normality of the distributions. The percentage of subjects (over the entire dataset) for which the null hypothesis was rejected, i.e., when the two distributions were significantly different, was reported. We refer to this metric as Precision Difference Percentage (PDP).
\label{sec:methods}

\subsection{Results}
The segmentation model achieved high accuracy for all the regions of interest in both datasets, with an average Dice score of 87\% across all segmented structures at both ED and ES.

Table \ref{tab:mean_and_std} shows the mean and standard deviation of the mean differences between the scan and rescan biomarker distributions, and the CoV between scan A and B for each biomarker. For both datasets, the lowest mean and standard deviation values were obtained from MC dropout, indicating good consistency between biomarker values extracted from scan A and B. The two other methods obtained mean difference values similar to MC dropout, but with higher standard deviations, reflecting the higher variability in the segmentation outcomes obtained using DE and TTA. In terms of CoV, all methods resulted in similar performance, with low values of the metric suggesting high precision. Overall, based on these metrics, all methods achieved satisfactory levels of precision.

\begin{table}[h]
\centering
\resizebox{\columnwidth}{!}{ %
\begin{tabular}{l|c|c|c|c|c|c}
\hline\hline
 \multicolumn{7}{c}{\textbf{Mean $\pm$ standard deviation of differences and CoV values} ($\downarrow$)} \\
\hline\hline
 & \multicolumn{6}{c}{\textbf{Roca-HC}} \\
 \hline
 & \multicolumn{2}{c|}{DE} & \multicolumn{2}{c|}{TTA} & \multicolumn{2}{c}{MC-D}\\
\hline
 & Mean $\pm$ std & CoV & Mean $\pm$ std & CoV & Mean $\pm$ std & CoV \\
\hline
LVEF & 3.21 $\pm$ 0.62 & \textbf{2.52} & 3.20 $\pm$ 0.77 & \textbf{2.52} & \textbf{3.11 $\pm$ 0.26} & \textbf{2.52} \\
LVM & 3.58 $\pm$ 0.90 & \textbf{1.74} & 3.81 $\pm$ 1.37 & 1.84 & \textbf{3.58 $\pm$ 0.38} & 1.84 \\
RVEF & 4.01 $\pm$ 0.99 & \textbf{3.38} & 4.62 $\pm$ 1.99 & 3.90 & \textbf{ 4.15 $\pm$ 0.46} & 3.64\\
\hline 
& \multicolumn{6}{c}{\textbf{Strain-8}} \\
 \hline
 & \multicolumn{2}{c|}{DE} & \multicolumn{2}{c|}{TTA} & \multicolumn{2}{c}{MC-D}\\
\hline
 & Mean $\pm$ std & CoV & Mean $\pm$ std & CoV & Mean $\pm$ std & CoV \\
 \hline
LVEF & 2.78 $\pm$ 0.45 & 2.20 & 3.11 $\pm$ 0.65 & 2.48 & \textbf{2.73 $\pm$ 0.16} & \textbf{2.20} \\
LVM & 3.04 $\pm$ 0.77 & 1.56 & 3.02 $\pm$ 1.13 & \textbf{1.55} & \textbf{2.98 $\pm$ 0.32} & 1.61 \\
RVEF & 2.75 $\pm$ 0.74 & 2.42 & 4.00 $\pm$ 1.89 & 3.78 & \textbf{2.55 $\pm$ 0.39} & \textbf{2.33} \\
\hline
\end{tabular}
} %
\caption{Mean and standard deviation of the mean differences between scan and rescan biomarker distributions and CoV between scan A and scan B}
\label{tab:mean_and_std}
\end{table}

\begin{figure*}[t]
    \centering
    \includegraphics[width=\textwidth]{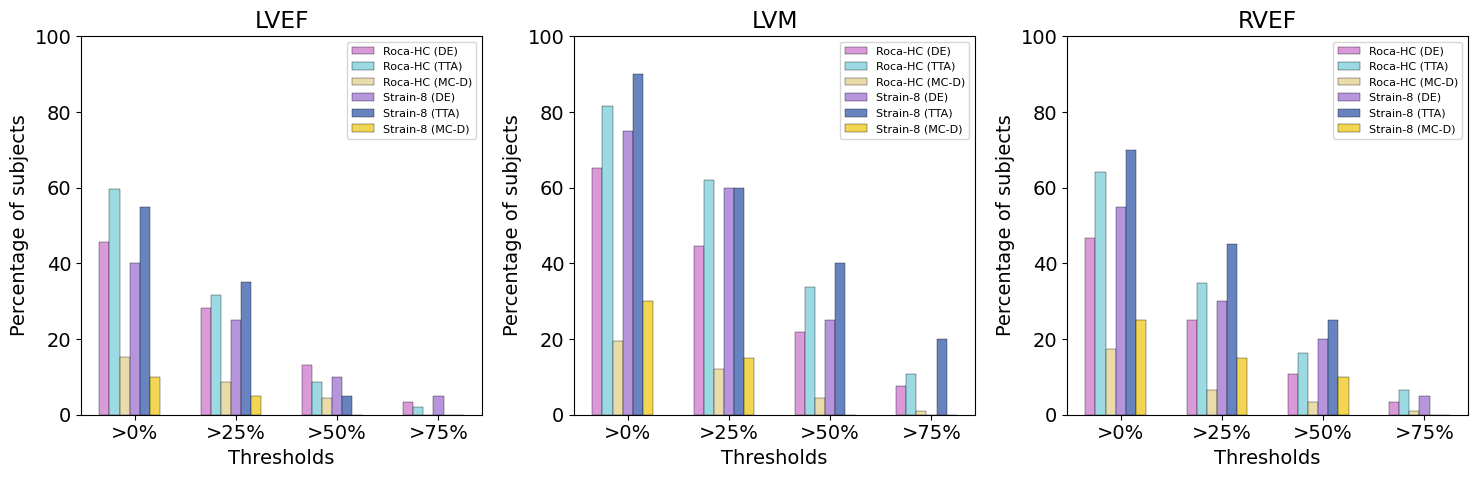}
    \caption{Distribution of subjects according to CIoU values}
    \label{fig:overlaps}
\end{figure*}

The agreement between the CIs of the scan and rescan biomarker distributions are shown in Table \ref{tab:percentages_ci} (CPP) and Figure \ref{fig:overlaps} (CIoU). While DE and TTA obtained similar CPP values, the performance of MC dropout was significantly worse. Similarly, the bar charts in Figure \ref{fig:overlaps}, which show the percentage of subjects with CIoU values which are $>$0\%, $>$25\%, $>$50\%, and $>$75\%, show low values of CIoU for MC dropout across the datasets. These results demonstrate that the agreement in the CIs between the scan and rescan distributions was low, in contrast with the results in Table \ref{tab:mean_and_std}.

\begin{table}[h]
\resizebox{\columnwidth}{!}{ %
\begin{tabular}{l|c|c|c|c|c|c}
\hline\hline
\multicolumn{7}{c}{\textbf{Confidence Precision Percentage (CPP)} ($\uparrow$)} \\
\hline\hline
 & \multicolumn{6}{c}{\textbf{Roca-HC}} \\ 
 \hline
 & \multicolumn{2}{c|}{DE} & \multicolumn{2}{c|}{TTA} & \multicolumn{2}{c}{MC-D} \\
  \hline
 & A $\rightarrow$ B & B $\rightarrow$ A & A $\rightarrow$ B & B $\rightarrow$ A & A $\rightarrow$ B & B $\rightarrow$ A \\ 
 \hline
LVEF & 31.52 & 28.26 & 27.17 & \textbf{33.70} & 6.52 & 7.61 \\
LVM & 39.13 & 42.39 & 52.17 & \textbf{57.61} & 11.96 & 9.78 \\
RVEF & 28.26 & 26.09 & 35.87 & \textbf{36.96} & 7.61 & 9.78 \\
\hline
 & \multicolumn{6}{c}{\textbf{Strain-8}} \\
 \hline
 & \multicolumn{2}{c|}{DE} & \multicolumn{2}{c|}{TTA} & \multicolumn{2}{c}{MC-D} \\
  \hline
 & A $\rightarrow$ B & B $\rightarrow$ A & A $\rightarrow$ B & B $\rightarrow$ A & A $\rightarrow$ B & B $\rightarrow$ A \\ 
 \hline
LVEF & 20.00 & 30.00 & 25.00 & \textbf{35.00} & 10.00 & 5.00 \\
LVM & 40.00 & 55.00 & \textbf{60.00} & 50.00 & 10.00 & 10.00 \\
RVEF & 30.00 & 30.00 & \textbf{50.00} & 45.00 & 10.00 & 25.00 \\ 
\hline
\end{tabular}
} %
\caption{CPP values - Inclusion rate of the mean from distribution A in the CI from B, and vice versa}
\label{tab:percentages_ci}
\end{table}

The values of PDP shown in Table \ref{tab:similarity} show a high percentage of statistically significantly different scan-rescan biomarker distributions. The lowest PDP was obtained for DE when estimating RVEF in the Strain-8 dataset (65\%). In the case of MC dropout, more than 90\% of the cases resulted in significantly different distributions of scan and rescan biomarkers in both datasets.

\begin{table}[h]
\centering
\resizebox{\columnwidth}{!}{ %
\begin{tabular}{l|c|c|c|c|c|c}
\hline\hline
\multicolumn{7}{c}{\textbf{Precision Difference Percentage (PDP)} ($\downarrow$)} \\
\hline\hline
 & \multicolumn{3}{c|}{\textbf{Roca-HC}} & \multicolumn{3}{c}{\textbf{Strain-8}} \\
 \hline
& DE & TTA & MC-D & DE & TTA & MC-D \\
\hline
LVEF & \textbf{80.43} & 86.96 & 95.65 & 85.00 & 85.00 & 95.00 \\
LVM & \textbf{68.48} & 72.83 & 93.48 & 75.00 & 75.00 & 100.00 \\
RVEF & 73.91 & 75.00 & 95.65 & \textbf{65.00} & 90.00 & 90.00 \\
\hline
\end{tabular}
} %
\caption{PDP values - Rejection rate of the null hypothesis in statistical tests for similarity between scan and rescan biomarker distributions}
\label{tab:similarity}
\end{table}

\begin{figure}[t]
    \centering
    \includegraphics[width=\columnwidth]{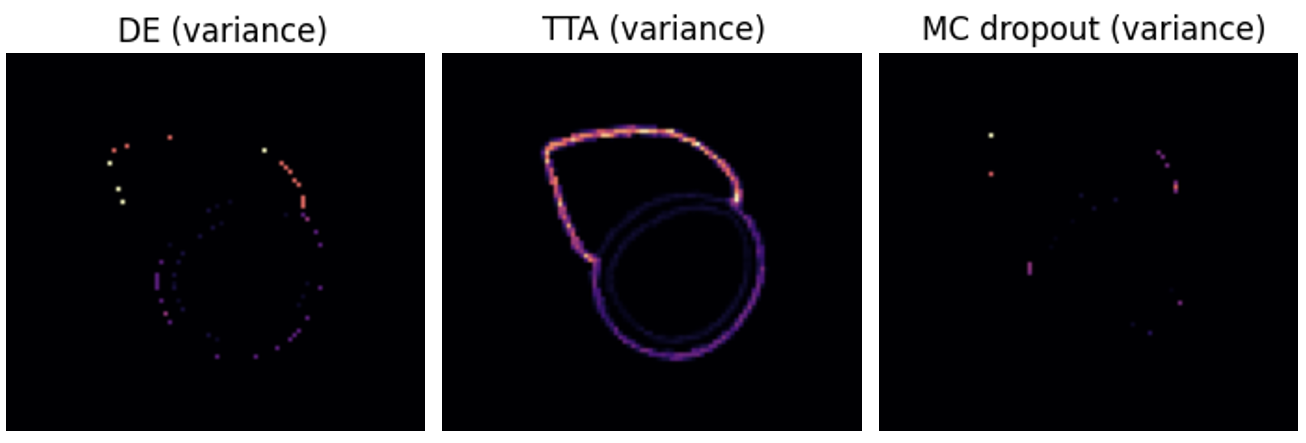}
    \caption{Uncertainty maps generated for a single case for each method}
    \label{fig:uncertainty_maps}
\end{figure}

Figure \ref{fig:uncertainty_maps} shows uncertainty maps obtained for a single case across the three employed methods. The variance is low for all methods, with most uncertainty being localised at the edges of the segmented structures.
 
\label{sec:results}

\section{Discussion and Conclusions}
This study used uncertainty estimation techniques to derive multiple biomarker estimates from cine CMR segmentations, and used the derived biomarker distributions to quantify the precision of the employed segmentation model on scan-rescan datasets. 

The mean and standard deviation of the scan-rescan differences and the CoV values demonstrated good levels of agreement for all uncertainty estimation methods. However, from the analysis of the distribution-based metrics we propose in this paper, MC dropout showed significantly lower precision between the distributions of scan A and scan B biomarkers. This highlights that the aggregation of multiple biomarker estimates can result in a seemingly high precision between the scan and rescan outputs, however, the comparison of the CIs derived from the biomarker distributions provides a more stringent test of the differences between such distributions. This analysis demonstrates that aggregation metrics may be insufficient to measure the agreement between biomarkers derived from two scans belonging to the same patient and motivates the use of distribution-based metrics such as the ones proposed in this work.

Due to the short period of time occurring between the two acquisitions, the main source of variability in scan-rescan data is related to the replanning of the scan and repositioning of the patient. This variability is likely captured by the intensity and spatial transformations applied to the inputs with TTA. In fact, this method resulted in the highest agreement in terms of CIs, as showed in Table \ref{tab:percentages_ci} and Figure \ref{fig:overlaps}. DE obtained similar levels of precision to TTA, while MC dropout produced narrow CIs, with lower degrees of overlap between scan and rescan. These outcomes emphasise the importance of aleatoric uncertainty in determining the precision of biomarker estimates.

As seen in Figure \ref{fig:overlaps}, the CIoU between scan and rescan distribution is lower than 50\% for the majority of subjects. The outcomes of the statistical tests in Table \ref{tab:similarity} also demonstrate that the scan and rescan distributions of biomarkers are significantly different for most subjects. Ideally, scan and rescan biomarkers should come from the same distribution, however this is often not the case. This shows how distribution-based metrics are more representative for precision analyses and offer a more in depth understanding of the agreement between scan and rescan biomarker distributions compared to standard precision metrics such as mean and standard deviation, or CoV.

In conclusion, this work demonstrates the importance of assessing the precision of DL-based methods for CMR analysis. For this purpose, we propose new metrics to characterise the change in biomarker distributions obtained from scan and rescan acquisitions. These metrics may provide a better assessment of precision, and highlight the need for the implementation of techniques that aim to increase the repeatability of cardiac functional biomarker estimates.

\label{sec:discussion_conclusion}




\section{Compliance with ethical standards}
The protocol for the Roca-HC dataset received full ethical approval from South Central - Berkshire B Research Ethics Committee (20/SC/0185). Participants gave informed consent to participate in the study before taking part. The Strain-8 prospective study was approved by the institutional review board at Guy’s and St Thomas
Hospital National Health Service Foundation Trust and the United Kingdom Health Research
Authority. Participants provided informed written consent.

\label{sec:ethics}



\section{Acknowledgments}
We would like to acknowledge funding from the EPSRC Centre for Doctoral Training in Medical Imaging (EP/S022104/1) and the Wellcome/EPSRC Centre for Medical Engineering at King’s College London (WT 203148/Z/16/Z). We also acknowledge funding from Perspectum Ltd., Oxford, UK.
\label{sec:acknowledgments}




\bibliographystyle{IEEEbib}
\bibliography{strings,refs}
\end{document}